%% file: main.tex
\documentclass[conference]{IEEEtran}

\usepackage{cite}
\usepackage[numbers]{natbib}
\usepackage[nottoc]{tocbibind}

\ifCLASSINFOpdf
   \usepackage[pdftex]{graphicx}
   \graphicspath{{../pdf/}{../jpeg/}}
   \DeclareGraphicsExtensions{.pdf,.jpeg,.png}
\else
\fi

\usepackage{amsmath}
\usepackage{amsfonts}

\usepackage{array}

\usepackage{fixltx2e}

\usepackage{stfloats}

\usepackage{url}

\begin{document}

\setlength{\abovedisplayskip}{-7pt}
\setlength{\belowdisplayskip}{0pt}
\setlength{\abovedisplayshortskip}{-7pt}
\setlength{\belowdisplayshortskip}{0pt}


\title{Automated Diagnosis of Clinic Workflows\vspace{-0.2in}}


\author{\IEEEauthorblockN{Alex Cheng, Jules White
}
Vanderbilt University, \\
Nashville, TN \\
Email: \{alex.cheng,jules.white\}@vanderbilt.edu\vspace{-0.2in}}


%


\maketitle

\input{0-abstract}

\IEEEpeerreviewmaketitle

\vspace{-0.1in}
\input{1-introduction}

\input{2-motivation}

\input{4-csp}

\input{5-tech}

\input{6-results}

\input{7-discussion}

\input{7b-relatedWork}

\input{8-conclusion}

\vspace{-0.05in}
\section*{Acknowledgment}
\vspace{-0.05in}
The authors would like to thank the National Library of Medicine for supporting Alex Cheng's training grant.
\vspace{-0.05in}





\end{document}

%% file: 0-abstract.tex
\begin{abstract}
Outpatient clinics often run behind schedule due to patients who arrive late or appointments that run longer than expected. We sought to develop a generalizable method that would allow healthcare providers to diagnose problems in  workflow that disrupt the schedule on any given provider clinic day. We use a constraint optimization problem to identify the least number of appointment modifications that make the rest of the schedule run on-time. We apply this method to an outpatient clinic at Vanderbilt. For patient seen in this clinic between March 27, 2017 and April 21, 2017, long cycle times tended to affect the overall schedule more than late patients. Results from this workflow diagnosis method could be used to inform interventions to help clinics run smoothly, thus decreasing patient wait times and increasing provider utilization.

\end{abstract}

\begin{IEEEkeywords} 
mHealth, workflows, constraint satisfaction, optimization, diagnosis
\end{IEEEkeywords}

%% file: 1-introduction.tex


\section{Introduction}
\vspace{-0.05in}

With the high cost and competitive landscape of the healthcare industry~\cite{Bodenheimer2005}, health services researchers have applied operations research methods in an effort to decrease costs or increase revenue~\cite{brandeau2004operations}. Additionally, patient wait times have been linked to patient satisfaction and perception of the quality of care~\cite{Bleustein2014}, and are an outcome that operational improvements can address. One area of interest for solving these problems in healthcare is scheduling optimization for outpatient appointments and procedures~\cite{Gupta2008}. Studies have used mathematical programming models to optimize for desired outcomes such as utilization, throughput, and patient wait times~\cite{Cayirli2003}. Other studies have used stochastic models such as discrete event simulations to describe complex clinical processes~\cite{Jun2009}. These studies tune resource constraints such as staffing, equipment, or rooms to improve simulated outcomes~\cite{Mielczarek2012}. In healthcare, these methods are typically applied to busy and high value areas of the system such as chemotherapy~\cite{Le2015}, surgery~\cite{Cardoen2010}, radiation therapy~\cite{Saure2012}, or the emergency department~\cite{Hoot2008}.

\textbf{Open problem.} There are several problems with simulation and mathematical models developed in previous studies. First, models describing healthcare processes are specific to a clinic or institution, making the model difficult or impossible to generalize to other use cases~\cite{Roberts2011}. Additionally, these models are difficult to validate with workflow data. Finally, model variables such as procedure times are often multi-faceted or non-modifiable for clinical reasons, thus complicating interventions designed to improve workflow. While many studies have sought to optimize scheduling or resources in order to improve certain outcomes, little work has been done to automate the identification of problems with clinic operations given real-world data. 

\textbf{Key contributions.} Unlike previous studies that optimize for a given utility function or outcome, our study seeks simply to diagnose problems with clinic workflow that cause appointments to start later than scheduled. Our model makes no assumptions about resources or existing distributions of services times. 
Therefore, our model is generalizable to any care setting or institution where data is available for scheduled appointment time, scheduled appointment duration, actual patient arrival time, and actual appointment duration. 

This paper provides the following contributions to the study of computer aided clinic workflow diagnosis:
\begin{itemize}
\item It discusses how a constraint satisfaction model can depict the existing state of patient arrival times, appointment start times, and appointment durations.
\item It discusses how comparing the existing state to scheduled appointment times can show mismatches in the planned and actual schedules.
\item It discusses how a constraint optimization problem can diagnose whether late patients, poor appointment duration allocation, or variability in treatment duration most likely led to the mismatch between planned and actual schedules.
\end{itemize}

%% file: 2-motivation.tex
\section{Motivating Example}
\label{motivation}
\vspace{-0.05in}
We apply our constraint satisfaction problem to  appointments at an outpatient clinic of Vanderbilt University Medical Center between March 27 and April 21, 2017. The basis for our actual schedule are timestamps for when the patient arrives at the clinic, when the patient moves to the exam room for the start of their appointment, and when the patient leaves the clinic. Timestamp data for patient flow are collected by two systems in that area. One system is a workflow management tool integrated with the electronic medical record, where staff track the progress of patients through their appointments\cite{Weinberg2006}. The second system is an automated patient tracking tool, where patients receive a Bluetooth low energy beacon that tracks their room location within the clinic. For each checkpoint in the patient process, we take the earlier timestamp of the two systems to improve accuracy.

We also pre-processed actual cycle times by assuming that the provider clinic is a single server process. This means that providers only saw one patient at a time in order of their appointment start times. Since most providers see patients in multiple rooms there are many cases where a patient room-in time overlaps with the next patient. In this case, we assume that the earlier patient departed and the later patient arrived in the room halfway through the time where their room-in times overlapped.

The planned schedule is taken from the appointment record. Each appointment has a scheduled start time and scheduled duration. The mismatch between the scheduled appointments and the timestamp data are the basis for our constraint satisfaction model.

%% file: 4-csp.tex
\vspace{-0.1in}
\section{Constraint Satisfaction Model of Patient Cycle Times}
\label{csp}
\vspace{-0.05in}

Constraint satisfaction problems (CSPs) are defined by a set of variables, such as the positive integer variables X and Y, and a set of constraints over the variables, such as $X < Y$. A constraint satisfaction problem defines a number of variables and constraints. A valid solution to a CSP is an assignment of values to the variables that adheres to all of the constraints. For example, X = 1, Y = 2, is a valid solution to this CSP. An assignment of values to the variables is called a labeling.

Constraint solvers are automated tools that are used to solve CSPs. A constraint solver takes a set of variables, constraints, and any initial labelings of variables as input. The solver then automatically produces valid labelings for the remaining unlabeled variables that satisfy the constraints. For example, if a constraint solver was provided the CSP above and an initial labeling of Y = 3, it would solve for the valid labelings of X, 1 and 2. 

In order to use a constraint solver, a CSP must first be defined that captures the relationships between the variables of interest. In this paper, the cycle times of patients and their appointment times are of interest. This section walks through the construction of an initial CSP that captures the relationship between planned appointment times and durations, and actual observed appointment times and durations. In Section~\ref{auto}, this CSP is extended in a way that allows a constraint solver to automatically derive answers to whether late patients or long cycle times are responsible for clinics running behind schedule.

Before beginning discussion of the model, a few key assumptions must be expressed. These key assumptions are outlined in Table~\ref{assumed}. The most important assumption is that we analyze the schedule for a single provider at a time. Analysis of multiple providers are possible, but each will have a separate CSP model built for their analysis.
  \vspace{-0.1in}
\begin{table}[h]
\caption{Key Model Assumptions}
\vspace{-0.15in}
\label{assumed}
\begin{center}
\begin{tabular}{|l|p{7cm}|}
\hline 
 A1. & A model is built for each individual provider's schedule and patients. \\
\hline
 A2. & A provider completes appointments sequentially. \\
 \hline
 A3. & The appointment times T and At are sorted in ascending order based
       on actual start time. \\
\hline
\end{tabular}
\end{center}
\vspace{-0.2in}
\end{table}

\begin{table}[h!]
\caption{CSP Workflow Variables}
\vspace{-0.15in}
\label{cspvars}
\begin{center}
\begin{tabular}{|l|p{3.5cm}|}
\hline 
 $T = \{T_0 \dots T_n\} \in [0, 1440] $ & Scheduled start time of appointment as minutes offset from midnight \\
\hline
 $D = \{D_0 \dots D_n\} \in [0, 1440] $ & Scheduled duration of appointment in minutes \\
\hline
 $At = \{At_0 \dots At_n\} \in [0, 1440] $ & Actual start time of appointment as minutes offset from midnight \\
\hline
 $As = \{As_0 \dots As_n\} \in [0, 1440] $ & Difference in minutes of scheduled vs. actual appointment start time \\ 
\hline
 $Ad = \{Ad_0 \dots Ad_n\} \in [0, 1440] $ & Actual duration of appointment in minutes \\ 
\hline
 $Ae = \{Ae_0 \dots Ae_n\} \in [0, 1440] $ & Difference in minutes of scheduled vs. actual appointment duration \\ 
\hline
 $Ap = \{Ap_0 \dots Ap_n\} \in [0, 1440] $ & Actual patient arrival time as minutes offset from midnight \\  
 \hline
 $F = \{F_0 \dots F_n\} \in [0, 1440] $ & Actual end time of appointment as minutes offset from midnight \\
\hline
 $C = \{C_0 \dots C_n\} \in [0, 1440] $ & Cycle time of each patient in minutes \\
\hline
 $W = \{W_0 \dots W_n\} \in [0, 1440] $ & Difference between scheduled and actual cycle time in minutes \\
\hline
\end{tabular}
\end{center}
\vspace{-0.3in}
\end{table}

We begin our model by defining a basic CSP. In the next subsection, we introduce additional variables into this CSP to support automated wait time diagnosis. The basic form of the CSP is shown in equations~\ref{eq1}-\ref{eq3}.
The CSP input to the constraint solver is composed of a planned or expected schedule, $E = <T,D>$, and a set of actual observed values, $A = <At, Ad>$. A cycle time can be calculated for each appointment using either the planned values $cycle(E)$ or the actual observed values $cycle(A)$.

\begin{equation} 
\label{eq1}
       As_i = 
        \begin{cases}
            i = 0 & \max (0, ~Ap_i - T_i) \\
            i > 0 & \max (0, ~At_{i-1} + Ad_{i-1} - T_i,~Ap_i - T_i)
        \end{cases}
\end{equation}

Equation~\ref{eq1} defines the basic constraint covering the calculation of the difference in minutes between the expected start time of an appointment and the actual start time. The first appointment of the day, $As_0$, will either start on time or will be delayed by the difference in minutes between the scheduled start time and the arrival time of the patient $\max (0, ~Ap_i - T_i)$. If the patient is late, $As_0$ will be a positive number of minutes that the patient was late to their appointment. For all other appointments, the start time deviation will either be a result of late patient arrival or the late completion of the the preceding appointment in At. Note, At is sorted based on actual appointment start time and not scheduled start time, which allows the analysis to consider deviations from the planned schedule.

As shown in Equation~\ref{eq3}, the model constrains the actual duration of the appointment, $Ad_i$, to be equal to the expected duration of the appointment plus the difference between the expected and actual duration, $Ae_i$. This constraint is important later when the modified CSP is formulated to diagnose workflow issues.

\begin{equation} \label{eq3}
\begin{split}
Ad_i & = D_i + Ae_i \\
\end{split}
\end{equation}

Next, the model constraints the actual end time of a patient's appointment, $F_i$, to be the actual start time plus the expected duration of the appointment, $D_i$, and difference in expected and actual deviation of the appointment, $Ae_i$. This constraint is shown in Equation~\ref{eq4}

\begin{equation} \label{eq4}
\begin{split}
F_i & = At_i + D_i + Ae_i
\end{split}
\end{equation}

A key input into the CSP model is the goals for patient cycle time. Ideally, patients should have a cycle time that matches their scheduled appointment duration.
However, in reality, a patient may arrive late or a prior appointment may run late causing the cycle time and scheduled appointment time not to match. The model defines cycle time as the difference between the arrival time of the patient, $Ap_i$, and the actual finish time of the appointment, $F_i$. This constraint is shown in Equation~\ref{eq5}

\begin{equation} \label{eq5}
\begin{split}
C_i & = F_i - Ap_i 
\end{split}
\end{equation}

The final component of the model is the defining a goal variable, which is that, ideally, the scheduled cycle time of the appointment should match the actual cycle time of the appointment, $C_i$. Although it might seem that it is preferable for the actual cycle time to be less than the scheduled cycle time, this indicates potential overestimation and waste in the schedule that could allow for more appointments. Thus, the ideal schedule has as little deviation from the planned vs. actual cycle time. This goal constraint is shown in Equation~\ref{eq6}.

\begin{equation} \label{eq6}
\begin{split}
W_i & = C_i - D_i 
\end{split}
\end{equation}

With this simple CSP formulation of the model, all that a clinic can do is check that the actual collected data meets the expected constraints. If the data does not meet the constraints, it indicates a potential error in the data collection process or difference in actual operation vs. assumptions of this model. The next section extends the CSP model to allow automated analysis of whether late patients or long cycle times are responsible for clinics running behind schedule.

%% file: 5-tech.tex
\section{Constraint-based Diagnosis of Patient Cycle Times}
\label{auto}
\vspace{-0.1in}
The overall goal of the diagnosis process is to explain why the planned cycle times for patients are longer or shorter than the actual observed cycle times. 
The automated diagnosis process relies on using a constraint solver to derive changes that could have been made to either the planned schedule or the actual observed schedule that would make the expected and actual cycle times more closely align. For example, the automated diagnosis process may state that had a specific patient arrived on time, the entire schedule for the day would have matched expectations. Alternatively, the automated diagnosis process might state that the actual duration of a single appointment was much longer than planned, indicating that treatment was more complicated than expected, and threw off the schedule. These are the types of outputs that the modified CSP will produce.

In order to support these types of diagnoses, the model needs to encode the concept of a "change" that could be made to the actual or planned schedule to make them more closely align. The diagnosis tries to find the fewest changes to the actual schedule that would lead to actual cycle times matching planned cycle times. In other words, what things could have gone differently that would have made planned and actual cycle time the same. Later, the section will discuss how the constraint solver reasons over these changes to diagnose clinic workflows, since there are often a large number of possible changes that could be made to rectify the mismatch between planned and actual schedules.

\subsection{CSP Model of Cycle Time Diagnosis}
  \vspace{-0.05in}
More formally, given a planned schedules $E$ and $A$, such that $cycle(E) != cycle(A)$, the diagnosis defines a new CSP that solves for the set of changes R to E and A, such that $cycle(changes(E,R)) = cycle(changes(A,R))$. That is, the output of the CSP is a set of modifications to E and A that will make their calculated cycle times for each appointment equal.

To support the concept of a potential "change", the CSP model needs two additional variables introduced to model $R=<\delta Ae, \delta Ap>$. An overview of these variables is shown in Table~\ref{divars}. 

First, the variable, $Ae_i$, is set to 1 by the solver if changing the duration of the $i_{th}$ appointment to match the planned duration would make the actual and planned cycle times more closely align. Second, the variable $Ap_i$ is a variable set to 1 by the solver if changing the patient's arrival time to match the start time of the appointment would make the actual and planned schedules match more closely. 
\vspace{-0.1in}

\begin{table}[h]
\caption{CSP Diagnosis Variables}
\vspace{-0.15in}
\label{divars}
\begin{center}
\begin{tabular}{|l|p{3.5cm}|}
\hline 
 $\delta Ae = \{\delta Ae_0 \dots \delta Ae_n\} \in [0, 1] $ & The difference in actual vs. scheduled treatment time of the $i_{th}$ appointment should be set to 0.\\ 
\hline
 $\delta Ap = \{\delta Ap_0 \dots \delta Ap_n\} \in [0, 1] $ & The $i_{th}$ patient's arrival time should be changed to the start time of the appointment. \\  
\hline
\end{tabular}
\end{center}
\end{table}
\vspace{-0.1in}

In order to use these variables, they must be incorporated into the CSP constraints. 
The $\delta Ap_i$ change variable is incorporated into the CSP in Equation~\ref{eq7}. The variable $RAp_i$ models the difference in planned appointment start time and patient arrival time. If the $\delta Ap_i$ is set to 1, it indicates that the patient arrival time should be set to the appointment time in order to more closely match scheduled and actual cycle times. By setting $\delta Ap_i$ to 1, it causes $RAp_i$ to equal the original planned start time of the appointment.

\begin{equation} \label{eq7}
       RAp_i = 
        \begin{cases}
            \delta Ap_i = 0 & Ap_i \\
            \delta Ap_i = 1 & T_i
        \end{cases}
\end{equation}

The $\delta Ae$ variable is incorporated into the constraints in Equation~\ref{eq8}. If $\delta Ae$ is set to 1, $RAd_i$ takes the value of the original planned duration. Otherwise, $RAd_i$ takes the actual duration of the appointment as its value.

\begin{equation} \label{eq8}
       RAd_i = 
        \begin{cases}
            \delta Ae_i = 0 & D_i + Ae_i \\
            \delta Ae_i = 1 & D_i
        \end{cases}
\end{equation}

Finally, in Equation~\ref{eq10}, the model ties the new change variables to the calculation of the difference in planned vs. actual start time of the appointment. The constraint is a modified version of Equation~\ref{eq1} that uses the $RAp_i$, $RAt_i$, and $RAd_i$ variables. For example, if $Ap_i \neq T_i$, but $\delta Ap_i = 1$, $RAp_i$ will equal 0, just as it would have if the patient had arrived on time.

\begin{equation} \label{eq10}
       RAt_i = 
        \begin{cases}
            i = 0     & RAp_i \\
            i > 0 & \max (0, ~RAt_{i-1} + RAd_{i-1}, RAp_i)
        \end{cases}
\end{equation}

\begin{equation} \label{eq11}
\begin{split}
(RAt_i + RAd_i) = (T_i + D_i + \epsilon)
\end{split}
\end{equation}

\subsection{Diagnosis as Optimization}
  \vspace{-0.06in}
Clearly, there are arbitrarily many changes that could be made to the planned and actual schedules that would cause their cycle times for appointments to be the same. Therefore, a mechanism is needed to express to the constraint solver how to rank possible changes and diagnose the difference between an expected and actual schedule. The mechanism that the model uses to rank possible sets of changes is to try to minimize the total number of changes made to either the planned schedule, E, or the actually observed schedule, A. 

That is, the constraint solver is asked to solve for a solution that minimizes the value of Equation~\ref{eq12}. The solver is trying to find the minimal set of patients that could have arrived on time and appointments that could have met their expected duration to make the overall cycle times of all appointments match in both planning and actuality.

\begin{equation} \label{eq12}
\begin{split}
\sum_0^n \delta Ae_i + \delta Ap_i
\end{split}
\end{equation}

The output from the constraint solver will be a labeling of the variables in the CSP that minimizes the number of changes that have to be made to the planned or actual schedule to make them consistent. A key question is how this variable labeling can be used to answer questions about patient cycle times. The variables $\delta Ae_i$ and $\delta Ap_i$ are the path to answering these questions.

\subsubsection{Diagnosing: Are late patients responsible?} 

The $\delta Ap_i$ variables determine if the minimal set of changes to make the actual and planned cycle times align includes changing the arrival time of patients. If late patients are part of the minimal set of changes that can be used to explain the difference between planned and actual execution time, it indicates that late patients are a factor and can precisely pinpoint which patients contributed to throwing off the planned cycle times.

For example, if the 2nd patient's $Ap_2$ variable is set to 1 and there are not other outputs, it indicates that the solver can explain the discrepancy between the planned and actual cycle times simply by that patient's tardiness. Had that single patient arrived on time, actual cycle times for all appointments would have met their planned cycle times. The solver can output a single patient late arrival, multiple late arrivals, or a combination of late arrivals and poorly predicted appointment durations as the root cause.

If a large number or all arrival times of patients are suggested as needing to be changed, meaning most people are late, this is a potential indicator that the front desk check-in process is slow. Moreover, it could also indicate problems with accessibility of the clinic location, such as difficulty in finding parking or navigating to the clinic.

\subsubsection{Diagnosing: Are poor appointment block time estimates responsible?}

The $\delta Ae_i$ variables indicate that appointment treatment times explain the discrepancy between planned and actual cycle times. For example, if $Ae_3 = 1$, it indicates that the 3rd appointment of the day went over its expected duration and contributed to the discrepancy in planned and actual times. The solver can output a single or combination of appointments and late arrivals that created the issue.

If a large percentage of appointment durations exceeded expectations, meaning a large number of $\delta Ae_i$ variables are 1, it indicates a more pervasive issue with appointment block time planning. That is, if appointments are consistently over time, then it is likely that the provider is being scheduled insufficient time to see and treat each patient. Alternatively, if a single provider consistently has appointments that run past their expected duration, it may be that the particular provider slower or takes more time talking to their patients.

\subsubsection{Diagnosing: Is treatment time unpredictability responsible?}

If a single or small number of $\delta Ae_i$ variables consistently explain the discrepancy, it means that each day a small number of unpredictable appointments are running late and causing delays. For example, providers in urgent care clinics may face highly unpredictable health situations compared to other clinics with less emergent and varying conditions. Small numbers of $\delta Ae_i$ variables set to 1 indicate that it is unlikely that a clinic could have done anything differently to be on time.

\vspace{-0.05in}

%% file: 6-results.tex
\section{Results}
\label{results}
\vspace{-0.05in}

From March 27, 2017 to April 21, 2017, 14 providers saw at least five patients on at least one appointment day at the Vanderbilt University Medical Center clinic in our study. These providers completed a total of 622 appointments over this period. Out of all appointments, 116 started late due to the patient arriving after the scheduled time, while 256 ended after the allocated time due to delayed cycle times. 

Figure 1 shows an example of one provider's schedule on one day where a combination of late patients and long cycle times caused the clinic to run off schedule. In this example, the solver determined that the making the 9th patient on time and completing the 6th, 7th, 10th, 11th, 12th, 14th, 15th, and 16th appointments on schedule would cause the rest of the appointments to run on schedule.

In Table IV, we aggregate the diagnostic variables $\delta Ap$ and $\delta Ae$ for each provider over all their clinic days. Providers are sorted by the total number of patients seen. Provider A saw the most patients over the study period, and the solver determined that 12 patient check-in modifications and 40 appointment duration modifications were the minimum necessary to make that provider's clinics run on time. All providers had more appointment duration revisions than patient check-in revisions in the optimized schedule except for Provider D who had $\Sigma\delta Ap = 16 $ and $\Sigma\delta Ae = 14$, and Provider K who had $\Sigma\delta Ap = \Sigma\delta Ae = 5$.

\begin{figure*}
\centering
\includegraphics[width=\textwidth]{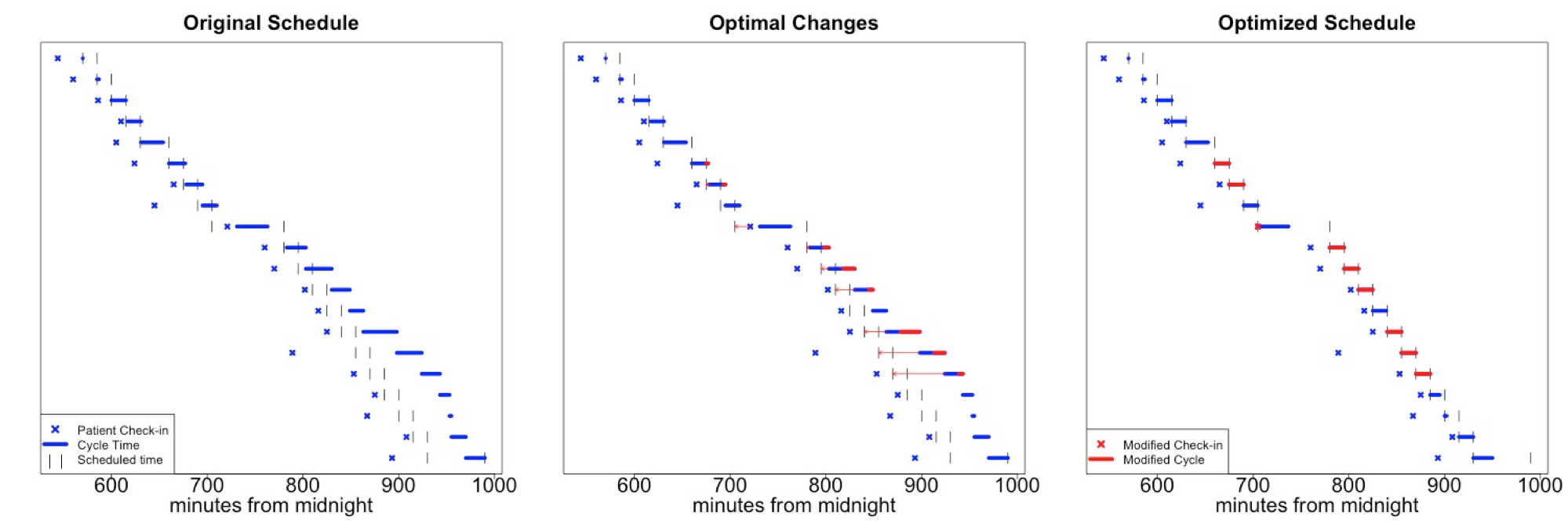}
\vspace{-0.35in}
\caption{Visualization of original and optimal patient check-in and cycle times} 
\label{fig:1}
\vspace{-0.25in}
\end{figure*}

\begin{table}[htbp]
  \centering
  \caption{Clinic Workflow Diagnosis by Provider}
  \vspace{-0.1in}
    \begin{tabular}{|l|r|r|r|r|}
    \hline
    Provider & $\Sigma \delta Ap $ & $\Sigma \delta Ae$ & Clinic Days & Patients Seen \\
    \hline
    A     & 12    & 40    & 7     & 106 \\
    B     & 5     & 11    & 7     & 66 \\
    C     & 15    & 22    & 4     & 66 \\
    D     & 16    & 14    & 10    & 63 \\
    E     & 12    & 27    & 4     & 59 \\
    F     & 5     & 6     & 8     & 51 \\
    G     & 7     & 20    & 2     & 45 \\
    H     & 11    & 13    & 4     & 40 \\
    I     & 12    & 12    & 6     & 38 \\
    J     & 12    & 15    & 3     & 38 \\
    K     & 5     & 5     & 3     & 24 \\
    L     & 2     & 7     & 3     & 15 \\
    M     & 0     & 0     & 1     & 6 \\
    N     & 2     & 2     & 1     & 5 \\
    \hline
    \end{tabular}%
  \label{tab:addlabel}%
\end{table}%

Table V shows aggregate totals for $\delta Ap$ and $\delta Ae$ by date across all providers who had clinic that day. Again, $\Sigma\delta Ae > \Sigma\delta Ap$ on most days except on March 29th, April 4th, April 19, and April 21. There does not appear, from our sample, to be any correlation between the ratio of $\Sigma\delta Ap : \Sigma\delta Ae$ and the number of patients seen, the number of providers, or the day of the week.

\begin{table}[htbp]
  \centering
  \caption{Clinic Workflow Diagnosis by Date}
  \vspace{-0.1in}
    \begin{tabular}{|r|r|r|r|r|}
    \hline
    Date & $\Sigma \delta Ap $ & $\Sigma \delta Ae$ & Patients Seen & \# Providers \\
    \hline
    27-Mar & 8     & 15    & 53    & 4 \\
    28-Mar & 4     & 15    & 41    & 4 \\
    29-Mar & 9     & 9     & 35    & 4 \\
    30-Mar & 8     & 14    & 41    & 4 \\
    31-Mar & 1     & 4     & 12    & 2 \\
    3-Apr & 7     & 13    & 48    & 4 \\
    4-Apr & 9     & 8     & 23    & 3 \\
    5-Apr & 6     & 11    & 30    & 3 \\
    6-Apr & 14    & 15    & 48    & 5 \\
    7-Apr & 3     & 6     & 18    & 3 \\
    10-Apr & 7     & 15    & 43    & 3 \\
    11-Apr & 9     & 14    & 46    & 4 \\
    12-Apr & 9     & 10    & 38    & 4 \\
    13-Apr & 6     & 14    & 38    & 4 \\
    14-Apr & 0     & 1     & 5     & 1 \\
    17-Apr & 3     & 5     & 20    & 2 \\
    18-Apr & 4     & 7     & 22    & 3 \\
    19-Apr & 6     & 6     & 26    & 3 \\
    20-Apr & 1     & 10    & 29    & 2 \\
    21-Apr & 2     & 2     & 6     & 1 \\
    \hline
    \end{tabular}%
  \label{tab:addlabel}%
\vspace{-0.25in}
\end{table}%

Finally, we aggregated $\delta Ae$ and $\delta Ap$ by the corresponding position of the revised appointment in the schedule in Table VI. For each provider clinic day with n appointments, any $\delta Ae$ and $\delta Ap$ in the first n/2 appointments (rounding down) would be assigned to the "first half" while the remainder would be assigned to the "second half". This means that provider clinic days with an odd number of appointments would have one more appointment attributed to the second half. Even with the discrepancy in the number of appointments favoring the second half, there were more modifications made to check-ins and cycle times in the first half of the schedule.

  \vspace{-0.1in}
\begin{table}[htbp]
  \centering
  \caption{Clinic Workflow Diagnosis by Position in Schedule}
  \vspace{-0.1in}
    \begin{tabular}{|r|r|r|}
    \hline
     & $\Sigma \delta Ap $ & $\Sigma \delta Ae$ \\
    \hline
    First Half of Schedule  & 63   & 116 \\
    \hline
    Second Half of Schedule & 53   & 78  \\
    \hline
    \end{tabular}%
  \label{tab:addlabel}%
\end{table}%

%% file: 7-discussion.tex
\section{Discussion}
\label{discussion}
\vspace{-0.05in}

\textbf{Interpretation of results.}
Our results demonstrate how we can use a constraint optimization problem to diagnose problems with clinic workflow. In diagnosing whether late patients are responsible for clinic going off schedule, we observed that for certain providers (such as Provider D in Table IV) and certain clinic days (such as April 4th in Table V), changing the arrival times for late patients would have caused the rest of the day to run according to schedule more so than adjusting planned appointment duration. Providers where $\Sigma\delta Ap < \Sigma\delta Ae$ may benefit from better coordination with the patient before their appointment in the form of appointment reminders, driving directions, or valet parking. Similarly, if the clinic notices trends in days that lead to high $\Sigma\delta Ap$, administrators could send reminders to patients ahead of days where tardy patients are likely to make a large impact on the schedule.

From our study sample, we are able to diagnose that poor appointment block time estimates are largely responsible for planned schedule breakdown. For most providers and clinic days, a large number of changes to appointment duration are needed to make the clinic run on schedule. This finding implies that there is overscheduling of patients where planned appointment time allocation is insufficient to address patient needs. The identification of these challenges could lead the clinic to make changes to clinic operations such as increasing planned appointment times, extending clinic hours, or increasing the number of providers.

Finally, we observe in Table IV that the solver made more schedule optimization changes in the first half of provider clinic days. This result is not surprising since a late patient or longer than expected appointment  early in the day can adversely affect the rest of the schedule. This finding may lead providers to schedule fewer patients and longer appointment blocks in the first half of the day to increase the likelihood of later appointments running on time.

\textbf{Current Limitations.}
Despite the effectiveness of this model in identifying problems with clinic workflow, there are several limitations that affect the validity and generalizability of this work. Firstly, our model does not account for interaction between potential changes and other appointments. By keeping $RAd_i = D_i$ where $\delta Ae_i = 0$ we assume that providers do not adjust the time they spend with patients based on their workload. In fact, providers may speed up or slow down their encounters with patients based on whether or not they are behind schedule. Another limitation of our model is that treating clinic operation as a single server process may be an oversimplification. Once patients enter exam rooms, they are often seen by multiple healthcare professionals. 
Finally, we assume in the constraint optimization problem that the least number of changes $\delta Ap$ and $\delta Ae$ is the best for getting the clinic back on schedule, even though some interventions may be easier to implement than others. 

%% file: 7b-relatedWork.tex
\vspace{-0.05in}
\section{Related Work}
\label{Related Work}
\vspace{-0.08in}
While this work is the first to use a CSP to diagnose problems in clinic workflow, other studies have used CSPs to create schedules in healthcare settings. Healthcare organizations use CSPs to solve nurse scheduling problems where a program will create a staffing schedule that satisfies hard constraints such as 24-hour coverage for inpatient units, while optimizing for soft constraints such as nurse preference \cite{Cheang2003}.

In non-healthcare domains, application developers have used constraint satisfaction optimization to identify the least number of software and hardware feature changes necessary to satisfy a set of dependency constraints\cite{White2010}. Similarly to this study, the CSP was used to identify conflicts in the existing feature sets, while the aggregated optimal number of modifications allowed developers to diagnose the design elements that needed the most work.

%% file: 8-conclusion.tex
\vspace{-0.05in}
\section{Conclusions}
\vspace{-0.08in}
\label{conclusion}

The results from this constraint optimization problem offer valuable insights that could help improve workflow in outpatient settings. The minimum number of changes to patient check-in times and appointment durations reveal whether patients or the healthcare system are responsible for the clinic running behind schedule. Using this method to diagnose previous clinic schedules can inform interventions that decrease patient wait times and improve provider utilization.

%% file: main.bbl
\begin{thebibliography}{10}

\bibitem{Bodenheimer2005}
Thomas Bodenheimer.
\newblock {High and Rising Health Care Costs. Part 1: Seeking an Explanation}.
\newblock {\em Annals of Internal Medicine}, 142(10):847, may 2005.

\bibitem{brandeau2004operations}
Margaret~L Brandeau, Fran{\c{c}}ois Sainfort, and William~P Pierskalla.
\newblock {\em Operations research and health care: a handbook of methods and
  applications}, volume~70.
\newblock Springer Science \& Business Media, 2004.

\bibitem{Bleustein2014}
Clifford Bleustein, David~B Rothschild, Andrew Valen, Eduardas Valatis, Laura
  Schweitzer, and Raleigh Jones.
\newblock {Wait times, patient satisfaction scores, and the perception of
  care.}
\newblock {\em The American journal of managed care}, 20(5):393--400, may 2014.

\bibitem{Gupta2008}
Diwakar Gupta and Brian Denton.
\newblock {Appointment scheduling in health care: Challenges and
  opportunities}.
\newblock {\em IIE Transactions}, 40(9):800--819, jul 2008.

\bibitem{Cayirli2003}
Tugba Cayirli, Emre Veral, and Inform Global.
\newblock {Outpatient Scheduling in Health Care: a Review of Literature}.
\newblock {\em Production and Operations Management}, 12(4):519--549, 2003.

\bibitem{Jun2009}
JB~Jun, SH~Jacobson, and JR~Swisher.
\newblock {Application of discrete - event simulation in health care clinics :
  A survey}.
\newblock {\em Journal of the Operational Research Society}, 50:109--123, 2009.

\bibitem{Mielczarek2012}
Bo{\.{z}}ena Mielczarek and Justyna Uzia{\l}ko-Mydlikowska.
\newblock {Application of computer simulation modeling in the health care
  sector: A survey}.
\newblock {\em Simulation}, 88(2):197--216, 2012.

\bibitem{Le2015}
Minh~Duc Le, Minh~H. {Nhat Nguyen}, Chantal Baril, Viviane Gascon, and Tien~Ba
  Dinh.
\newblock {Heuristics to solve appointment scheduling in chemotherapy}.
\newblock {\em Proceedings - 2015 IEEE RIVF International Conference on
  Computing and Communication Technologies: Research, Innovation, and Vision
  for Future, IEEE RIVF 2015}, pages 59--64, 2015.

\bibitem{Cardoen2010}
Brecht Cardoen, Erik Demeulemeester, and Jeroen Beli{\"{e}}n.
\newblock {Operating room planning and scheduling: A literature review}.
\newblock {\em European Journal of Operational Research}, 201(3):921--932,
  2010.

\bibitem{Saure2012}
Antoine Saur{\'{e}}, Jonathan Patrick, Scott Tyldesley, and Martin~L. Puterman.
\newblock {Dynamic multi-appointment patient scheduling for radiation therapy}.
\newblock {\em European Journal of Operational Research}, 223(2):573--584,
  2012.

\bibitem{Hoot2008}
Nathan~R. Hoot, Larry~J. LeBlanc, Ian Jones, Scott~R. Levin, Chuan Zhou,
  Cynthia~S. Gadd, and Dominik Aronsky.
\newblock {Forecasting Emergency Department Crowding: A Discrete Event
  Simulation}.
\newblock {\em Annals of Emergency Medicine}, 52(2):116--125, aug 2008.

\bibitem{Roberts2011}
Stephen~D. Roberts.
\newblock {Tutorial on the Simulation of Healthcare Systems}.
\newblock {\em Proceedings of the 2011 Winter Simulation Conference}, pages
  1408--1419, 2011.

\bibitem{Weinberg2006}
Stuart~T Weinberg, Dario~A Giuse, Randolph~A Miller, and Mark~A Arrieta.
\newblock {The Outpatient Clinic Whiteboard – Integrating Existing Scheduling
  and EMR Systems to Enhance Clinic Workflows}.
\newblock {\em AMIA Symposium Proceedings 2006}, 2006.

\bibitem{Cheang2003}
B~Cheang, H~Li, A~Lim, and B~Rodrigues.
\newblock {Nurse rostering problems––a bibliographic survey}.
\newblock {\em European Journal of Operational Research}, 151(3):447--460, dec
  2003.

\bibitem{White2010}
J.~White, D.~Benavides, D.C. Schmidt, P.~Trinidad, B.~Dougherty, and
  A.~Ruiz-Cortes.
\newblock {Automated diagnosis of feature model configurations}.
\newblock {\em Journal of Systems and Software}, 83(7):1094--1107, jul 2010.

\end{thebibliography}
